\documentclass[11pt]{article}
\usepackage{coling2018}
\usepackage{times}
\usepackage{url}
\usepackage{latexsym}

\usepackage{wrapfig}

\usepackage{subfig}

\usepackage{tabularx}
\usepackage{amssymb}
\usepackage{amsmath}
\usepackage{latexsym}
\usepackage{enumitem}
\usepackage{booktabs}
\usepackage{url}
\usepackage{color} 
\usepackage{verbatim}
\usepackage{calc}
\usepackage{arydshln}
\usepackage{mathtools}
\usepackage[draft]{todonotes}

\usepackage{soul}

\usepackage{mathtools}

\newtagform{brackets}{[}{]}
\usetagform{brackets}

\title{The Road to Success:\\Assessing the Fate of Linguistic Innovations in Online Communities}

\author{Marco Del Tredici \and Raquel Fern\'andez\\
  Institute for Logic, Language and Computation\\
  University of Amsterdam\\
  {\tt \{m.deltredici|raquel.fernandez\}@uva.nl}
}

\date{}

\begin{document}
\maketitle
\begin{abstract}
 We investigate the birth and diffusion of lexical innovations in a large dataset of online social communities.
We build on sociolinguistic theories and focus on the relation between the spread of a novel term and the social role of the  individuals  who use it, uncovering characteristics of innovators and adopters. 
Finally, we perform a prediction task that allows us to anticipate whether an innovation will successfully spread within a community.
\end{abstract}

\section{Introduction}
\label{sect:Introduction}


\blfootnote{
    %
  \hspace{-0.65cm}  
   This work is licenced under a Creative Commons 
   Attribution 4.0 International Licence.
   Licence details:
   \url{http://creativecommons.org/licenses/by/4.0/}
   }

Language is an incessantly evolving system, and linguistic innovations of different kind are continuously created. New variants originate in concrete communicative situations with particular speakers. After being introduced, some of them are adopted by other speakers belonging to the same community and possibly spread until they become community norms. In contrast, other innovations do not manage to make their way into the community conventions and just disappear after a certain period of time. This process raises many intriguing questions, which have been the focus of attention in Sociolinguistics since the late 1960's \cite{weinreich1968empirical,chambers2013handbook}. 
For example, at  what linguistic levels (phonology, lexicon, syntax) do innovations arise and succeed? 
Who are the leaders of change and what are their social characteristics? 
In this paper, we investigate lexical innovation in online communities, focusing on the latter type of question, by means of a large-scale data-driven approach. 

We study  the interplay between the birth and spread of new terms and users' social standing in large online social communities, taking as starting point hypotheses put forward in the Sociolinguistics literature (see Section~\ref{sec:related}). 
We present a longitudinal study on a large social media dataset, including 20 online forums and around 10 million users overall.
We consider each forum as an independent social network, and investigate how lexical innovations arise and spread within a forum. We analyse the fate of approximately 8 thousand innovations (focusing on acronyms such as \textit{lol}, phonetic spellings such as \textit{plis}, and other linguistic phenomena usually collected under the term of \textit{Internet slang}), and relate their spread within a social network to the role of the individuals who use them. We characterise users' roles within a community by means of a novel, theoretically-motivated \emph{tie-strength} measure, combined with well-known centrality measures. 
%

We show that (1) innovators (users who introduce a new term) are central members of a community, connected to many other users but with relatively low tie-strength, and (2) strong-tie users (who belong to cliques or sub-groups within the community) effectively contribute to the dissemination of a new term. This pattern is surprisingly consistent across the 20 online communities under investigation. In addition, we show that,
by solely using information on speakers' tie strength as predictor variable, we can anticipate whether an innovation will successfully spread within a community.

Our work yields new theoretical insights into the applicability of sociolinguistic theories to online settings. It also has practical significance for NLP systems encountering novel terms in social media:  
While new terms that do not succeed in becoming community norms may safely be treated as out-of-vocabulary words, a better understanding of the dissemination process of novel terms beyond their frequency can help identify those innovations that a system should be able to deal with. Together with the in-depth analysis presented in the paper, 
we make available our dataset, which includes interactions scraped from 20 topic-based forums over a period of 4 to 8 years, as well as the code for preprocessing the data, extracting the dissemination trajectories of $\sim$8k internet slang terms, and computing the strength of the ties among users.\footnote{\url{https://github.com/marcodel13/The-Road-to-Success}}



\section{Background}
\label{sec:related}


\subsection{Sociolinguistic Theories}

Our investigation of innovations in online social communities builds on Milroy's \shortcite{milroy1987language} theory, according to which linguistic innovations propagate and become community norms when individuals with {\em strong ties} adopt them early on; however, it is usually individuals with only \textit{weak ties} who are the actual {\em innovators}, i.e., those who introduce novel linguistic variants. Milroy's proposal is thus an application to Sociolinguistics of {\em The Strength of Weak Ties} theory put forward by \newcite{granovetter1973strength} to explain the diffusion of innovations more broadly.

In Milroy's \shortcite{milroy1987language} original study, the strength of a tie between two individuals is given by a combination of factors, such as the amount of time spent together, the level of intimacy, etc. As a result, strong ties are usually those connecting family members or close friends, while weak ties connect acquaintances. Individuals linked by strong ties form close-knit sub-communities or clusters, whereas weak ties act as bridges between these clusters. According to Milroy's theory, individuals without strong ties are more likely to innovate. Thanks to their bridging role, they can introduce an innovation into one or more close-knit clusters. Once introduced, the innovation can be adopted by closely  connected individuals within a cluster, who quickly propagate it to others.

Milroy's theory has found confirmation both in small-scale sociolinguistic studies carried out in the protestant enclaves of Belfast \cite{milroy1985linguistic,milroy1987belfast} and in agent-based models with simulated data~\cite{fagyal2010centers}. However, it has not been tested at a large scale on actual linguistic data.
%
Moreover, the theory contrasts with other well-established models regarding the source of novel variants, which casts some doubt on the generality of its application: Labov's studies on New York adolescent gangs and the neighbourhoods of Philadelphia \cite{labov1972,labov2001social} identified \textit{leaders}, i.e., those who have strong networks and are the most popular in their communities, as the drivers of change.
Similarly, Eckert's \shortcite{eckert2000language} studies of adolescent groups 
in Detroit showed that language novelty and change is led by charismatic leaders with strong ties to the local community.

We assess the applicability of Milroy's theory to the emergence of lexical innovations in online social communities, where relationships between individuals and their exposure to new terms can only be modelled in terms of their directly observable linguistic interaction in online discussion threads. 

\subsection{Related Computational Work}

Research on language change has recently experienced a boost within the NLP community \cite{hamilton2016diachronic,frermann2016bayesian}. Here we focus on reviewing approaches that are directly related to the work we present in this paper, i.e., approaches that investigate linguistic innovation within relatively short time spans of a few years (rather than looking at historical time) and that do so by leveraging data from social media to create a graph of connected users. More general literature on graph modelling and innovation diffusion is out of the scope of our review.

Graphs representing social networks have been used to analyse the extralinguistic factors that drive diffusion, such as geographical~\cite{eisenstein2015identifying} and demographic variables~\cite{eisenstein2014diffusion}, as well as the position of individuals in the social network~\cite{paolillo1999virtual} and the kind of user interactions that foster the diffusion of the innovations. For example, \newcite{goel2016social} and \newcite{paradowski2012diffusion} investigate the amount of interaction required to adopt an innovation, showing that while some innovations are adopted by a new user after multiple interactions, for others a single contact is sufficient. \newcite{rotabi2017competition} relate the success of competing lexical variants to the seniority of the users who use them. This work is extended by \newcite{rotabi2017tracing}, who introduce an inheritance graph to represent how speakers pass innovations (in this case called \textit{practices}) on to others in time through real-life collaborations.

Perhaps the work that is most directly connected to ours is that of~\newcite{rotabi2016status}, who investigate the diffusion patterns of words that experience a frequency burst at a given time, which the authors call {\em trends},
and analyse the level of activity of the users involved in the different phases of the diffusion process. While related, our work differs on key points: we focus on linguistic innovations rather than trends and characterise users in terms of tie strength rather than level of activity. 

A parallel line of work has investigated linguistic diffusion in relation to social ties using agent-based computer simulations \cite{ke2008language,fagyal2010centers,swarup2011model}. In particular, \newcite{fagyal2010centers} address questions similar to ours via simulated data and find that peripheral users are crucial in introducing new linguistic variants, which are then adopted and spread by central agents in the community. The current availability of large social media datasets containing real linguistic interactions allows us to study these questions with more ecological validity. 

\section{Methodology}
\label{sec:methods}


We describe our dataset, the methodology we use to define the social network of a community and the social role of its members, and the procedure to identify linguistic innovations and characterise their diffusion. 

\subsection{Data}

We use data from 
Reddit,\footnote{\url{https://www.reddit.com}} a popular website which includes around 1 million communities called {\em subreddits}. A subreddit is a topic-based forum where users can submit posts, 
comment on existing posts or score them. 
While in this paper we treat each forum independently (leaving the analysis of innovation diffusion across forums for future work),
in order to 
conduct a large-scale analysis and 
draw conclusions that generalise across communities, we analyse 20 different forums that show substantial variability in terms of subject matter and size (see Table~\ref{table:all_subreddits_stats} for an overview). Despite their rich heterogeneity, all subreddits we consider share the following features, which 
are indicative of highly active and interconnected communities, where the spreading cascade process at the base of innovation diffusion finds a favourable environment~\cite{hamilton2017loyalty,guille2013information}: 

\begin{itemize}[itemsep=0pt]
\item a topic that reflects a strong external interest, such as sport teams, videogames or TV series;
\item small-to-medium size, which in contrast to very large subreddits such as r/news (15M users) or r/funny (18M), are less dispersive and favour tighter connections;
\item high density, i.e., high ratio of existing connections over the number of potential connections;\footnote{\newcite{hamilton2017loyalty} report density values in the range [$0.001-0.016$] in their set of subreddits.}
\end{itemize}

\noindent
We downloaded the entire content of each subreddit from its first post to the end of 2016. 
We segment the data from each subreddit into consecutive time bins corresponding to one month.\footnote{We also experimented with smaller bins of one week, obtaining similar results.}
We discard time bins with less than 200 active users, which are common during the first few months of a subreddit lifespan, and ignore any posts whose author is unknown.\footnote{When users delete their account, 
the posts, comments, and messages submitted prior to the deletion are still visible to others, but information about the user is not available (see Reddit Privacy Policy at \url{https://www.redditinc.com/policies/privacy-policy).}} 

Next, we explain how we leverage this longitudinal data to extract information about the social role of forum members, as well as to detect linguistic innovations and characterise their diffusion.

\begin{table}\small
\centering
\begin{tabular}{lccccc@{\ }}\toprule
\bf subreddit & \bf years & \bf tokens & \bf users  & \bf density & {\bf innovations} \\\midrule

r/Android & 7 & 158 & 1.03M & 0.006 & 730\\
r/apple & 8 & 89 & 580k & 0.006  & 584\\
r/baseball & 6 & 101 & 576k & 0.014 & 520\\
r/beer & 7 & 29 & 291k & 0.008 & 360\\
r/boardgames & 6 & 88 & 313k & 0.004 & 380\\
r/cars & 6 & 101 & 544k & 0.014 & 605\\
r/FinalFantasy & 4 & 22 & 137k & 0.009 & 218\\
r/Guitar & 7 & 71 & 387k & 0.009 & 496\\
r/harrypotter & 5 & 39 & 287k & 0.005 & 227\\
r/hockey & 7 & 191 & 847k & 0.012 & 602\\
r/Liverpool & 5 & 40 & 173k & 0.018 & 314\\
r/Patriots & 5 & 26 & 151k & 0.009 & 231\\
r/pcgaming & 5 & 52 & 350k & 0.003 & 360\\
r/photography & 8 & 81 & 353k & 0.006 & 485\\
r/pokemon & 6 & 107 & 1.02M & 0.006 & 695\\
r/poker & 6 & 28 & 104k & 0.012 & 258\\
r/reddevils & 4 & 49 & 186k & 0.008 & 329\\
r/running & 6 & 56 & 279k & 0.008 & 367\\
r/StarWars & 6 & 56 & 542k & 0.008 & 381\\
r/subaru & 5 & 21 & 187k & 0.005 & 340\\
\bottomrule
\end{tabular}
\caption{Statistics of the subreddits in our dataset, including: years of activity considered until end of 2016; total \# of tokens (in millions); total \# of active users (including users who may have left the community); average ratio of network ties over all possible ties computed over all the time bins (density); total \# of linguistic innovations analysed.
\label{table:all_subreddits_stats}}
\end{table}

\subsection{Social Network}
\label{subsec:methods_social_network}



We create a graph representing a community's social network for each month $t$ during the community lifespan. In these graphs,  nodes are users and edges encode whether users have interacted.
We consider two users to be connected by an edge 
if they comment within the same thread in close proximity.\footnote{In particular, if they are separated by at most two posts, as done e.g.,  by \newcite{hamilton2017loyalty}.} Given that arguably lexical diffusion follows a simple ``contagion'' model whereby a single contact is sufficient for the spreading process~\cite{goel2016social}, our graphs are undirected and unweighted.\footnote{The graphs are implemented with Python's package \texttt{networkx}.}


Milroy's \shortcite{milroy1987language} theory relates the diffusion of a linguistic innovation to the local topology of individuals within a network, distinguishing between close-knit sub-groups and individuals outside of these groups. In line with other studies such as \newcite{weng2015attention} and \newcite{zhao2010weak}, we build on a measure introduced by \newcite{onnela2007structure}, which determines the strength of the edge between two individuals $i$ and $j$ in terms of the overlap ${O}_{ij}$ of their adjacent neighbourhoods, as follows:
\begin{equation}\label{eq:onnela}
{O}_{ij} = \frac{{n}_{ij}}{({k}_{i} - 1) + ({k}_{j} - 1) - {n}_{ij}}
\end{equation}

\noindent
where $n_{ij}$ is the number of adjacent nodes between $i$ and $j$, and $k_i, k_j$ their respective degree, i.e., the number of edges incident to each of them.
Possible values for $O_{ij}$ are in the range $[0,1]$, where $0$ indicates no common neighbours (weakest possible connection between $i$ and $j$) and $1$ exactly the same adjacent neighbours (strongest possible connection between $i$ and $j$). 




We now leverage Equation~[\ref{eq:onnela}] to characterize users given the strength of their connections.
According to \newcite{milroy1987language}, weak-tie individuals have only weak connections (are not part of close-knit clusters), while strong-tie individuals have strong connections with other users, but may also have weak connections if they are linked to weak-tie users. To capture this, we define the tie strength of each individual $i$ as the highest value of her incident edges. That is, for all individuals $j$ directly connected to $i$:
\begin{equation}\label{eq:tie-strength}
\emph{tie-strength}(i) := \max{(O_{ij})} 
\end{equation}

\noindent
Taking the maximum captures what we are after: A community member who only has weak connections will have low tie-strength and be considered a \emph{weak-tie user}, while a member with either strong connections only or with both strong and weak connections will have high tie-strength and be considered a \emph{strong-tie user}, who will be part of a local clique \cite{luce1949method}. 
Unlike mean or median values, which tend to be rather balanced and thus do not help to distinguish between innovators and non-innovators, taking the maximum is in line with Milroy's theory and captures the key difference between these two groups of users, as will be shown in the next section.

Besides computing tie-strength as defined in Equation~[\ref{eq:tie-strength}] for all users in our social graphs, we also compute their centrality values for three common measures of network centrality --- degree, betweenness, and eigenvector --- which are global indices of the importance of a node with respect to all other nodes in a graph \cite{newman2010networks}.

\subsection{Linguistic Innovations}



We focus on \textit{Internet slang}, a general term commonly used to refer to a range of linguistic phenomena such as abbreviations ({\em cu} for {\em see you}), acronyms ({\em IIRC} for {\em if I remember/recall correctly}) and phonetic spellings ({\em dat} for {\em that}). The motivation behind this choice is twofold: firstly, forms of this kind  are very abundant and continuously introduced in online communication; and secondly, they are easier to identify and track than innovations at other levels, such as those related to meaning shift. 
In the present study, we do not focus on tracking the co-evolution of two variants (e.g., {\em dat} vs.~{\em that}) that may be competing,  but rather on analysing the emergence of new forms and their trajectories independently, in light of the tie-strength of the members who use them.

Our starting point are the terms in the dictionary available at {\tt NoSlang.com},\footnote{\url{https://www.noslang.com/dictionary/}} 
 a comprehensive record of Internet slang that is constantly updated.
After removing terms including non-alphabetic characters, we obtain a list of approximately 6k terms.
For each subreddit, we only consider terms that: 

\begin{itemize}[itemsep=0pt]
\item are used at least 10 times in the subreddit; 
\item are not present during the first 3 months of the community's existence; and 
\item are introduced within the initial quarter of the community lifespan.
\end{itemize}

\noindent
That is, we restrict our analysis to \textit{newly} introduced Internet slang terms that are not present from the very beginning of the community's activity, but are not introduced too late so as to be able to observe their trajectory for a substantial period of time.\footnote{To assess whether ambiguity may be an issue, we checked whether the terms also appear in a standard English dictionary, PyDictionary. For example,  the slang term \emph{bra} for \emph{brother} also has the standard meaning of \emph{brasserie}. Given that under 2\% of terms in our dataset could potentially be ambiguous, we decided to not treat them in any special way.}

The number of innovations considered across all subreddits is 7962, while the number of unique innovations amounts to 1456. 
Most of the terms (around 76\%) occur in more than one community, although no innovation is present in all the subreddits in our dataset. 
Thus, around 24\% of innovations tracked occur in just one community --- some of these are clearly topic-related, e.g., {\em pkemon} in /r/pokemon, while others are more general purpose abbreviations that, in principle, could appear in any community, such as {\em txs} (\textit{thanks}, in /r/Android) or {\em omgz} (\textit{oh my god/gosh}, in /r/subaru).
Regarding frequency, while we set a minimum threshold of 10 occurrences, in practice 72\% of terms occur at least 50 times on average.


\paragraph{Dissemination trajectory.}
Once introduced, innovations may have different fates: they can spread widely within the community, be used by just a small sub-group, or fail to make an impact and disappear altogether. We define the fate of a term as its \textit{dissemination}, which we compute as the proportion of community members who use it at a given moment in time
\cite{deltredici-fernandez:2017:iwcs,altman-etal:plos-one:2011}. 
 It should be stressed that dissemination differs from frequency. Although often they are highly correlated, in principle a term can have high relative frequency but low dissemination and vice versa. We quantify the diffusion of innovations in terms of their dissemination since this gives us a measure of their spread within a community. To this end, for each innovation we calculate its {\em dissemination trajectory} as the vector of its monthly dissemination values since the innovation was introduced.  

\paragraph{Tie-strength trajectory.} 
Similarly, we compute the 
\textit{tie-strength trajectory} of each innovation as a vector whose features correspond to the maximum tie-strength value among the users that used it in the corresponding month. We choose the maximum value in order to test Milroy's \shortcite{milroy1987language} hypothesis, according to which the crucial factor in the diffusion of an innovation is its adoption by strong-tie members (see Section \ref{sec:related}).
Considering only the maximum value provides a simple way to test whether any individual with high tie-strength has used the term in a given month.\footnote{The dissemination and the tie-strength vectors always have the same magnitude.} 
\section{Empirical Observations}
\label{sec:empirical_observations}


\begin{figure*}[t]
\hspace*{-.33cm}
\begin{minipage}{5.12cm}
\includegraphics[height=4.5cm]{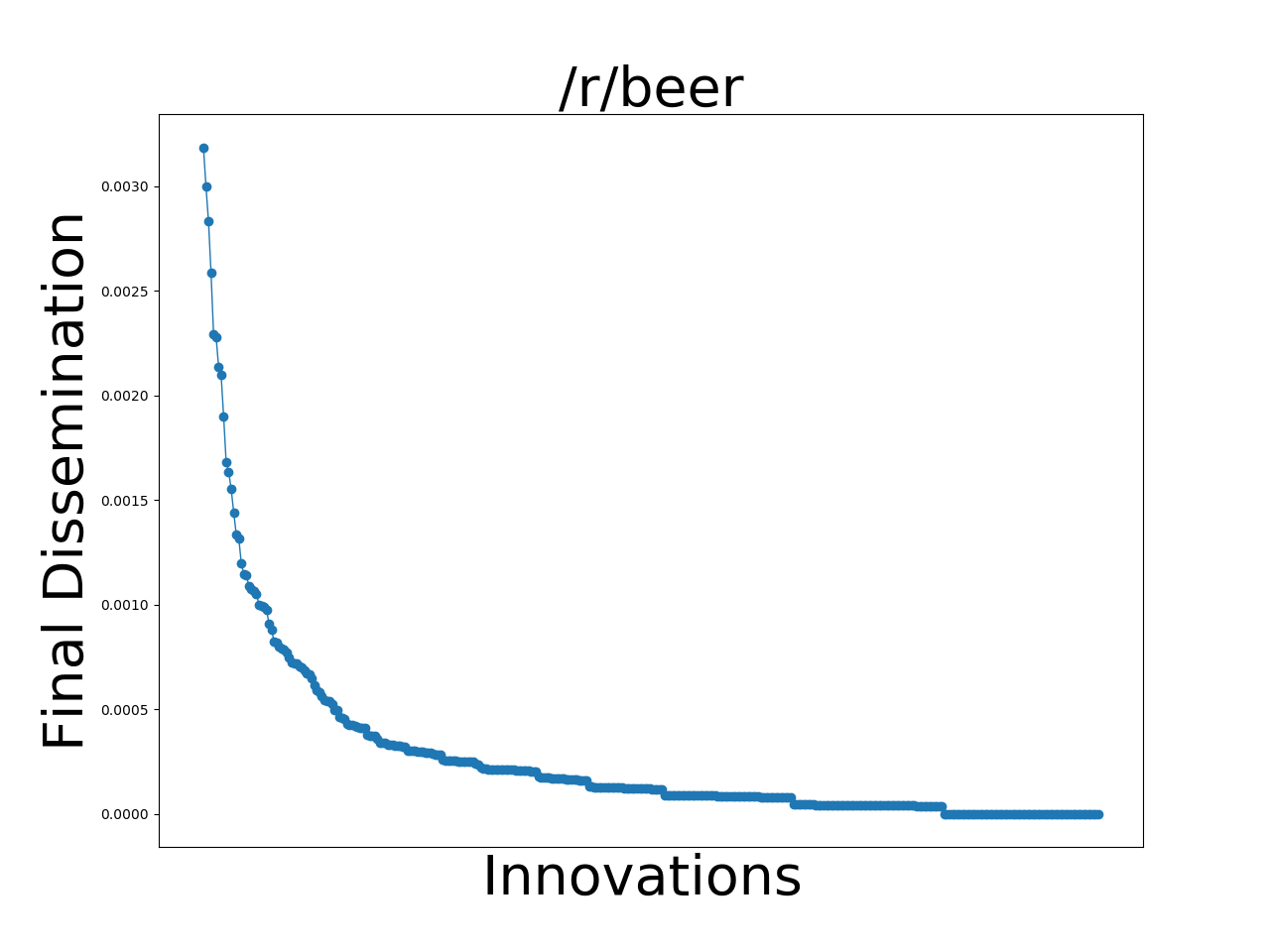}
\end{minipage} \ \ 
\begin{minipage}{5.35cm}
\includegraphics[height=4.5cm]{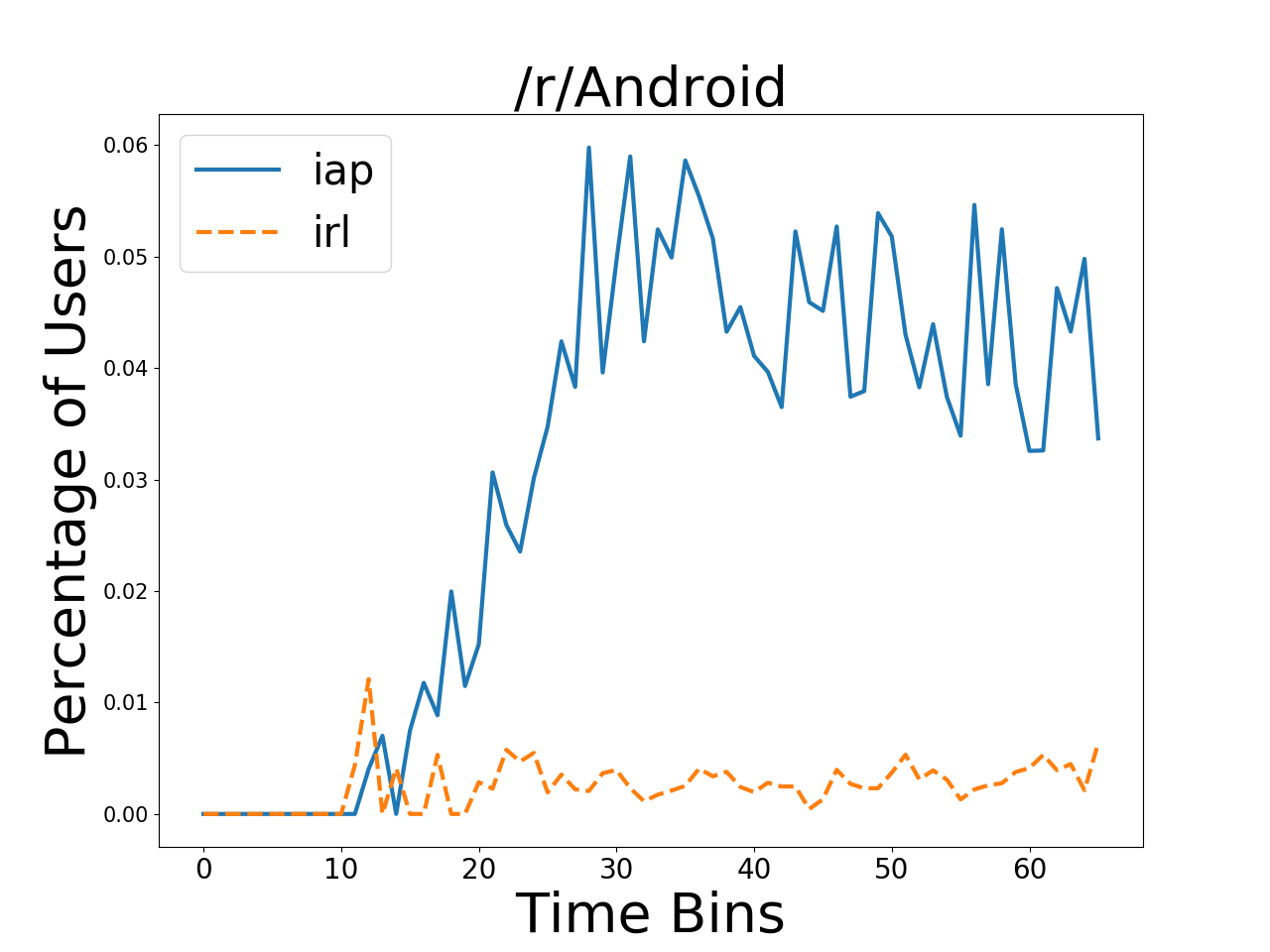}
\end{minipage}
\begin{minipage}{5.35cm}
\includegraphics[height=4.5cm]{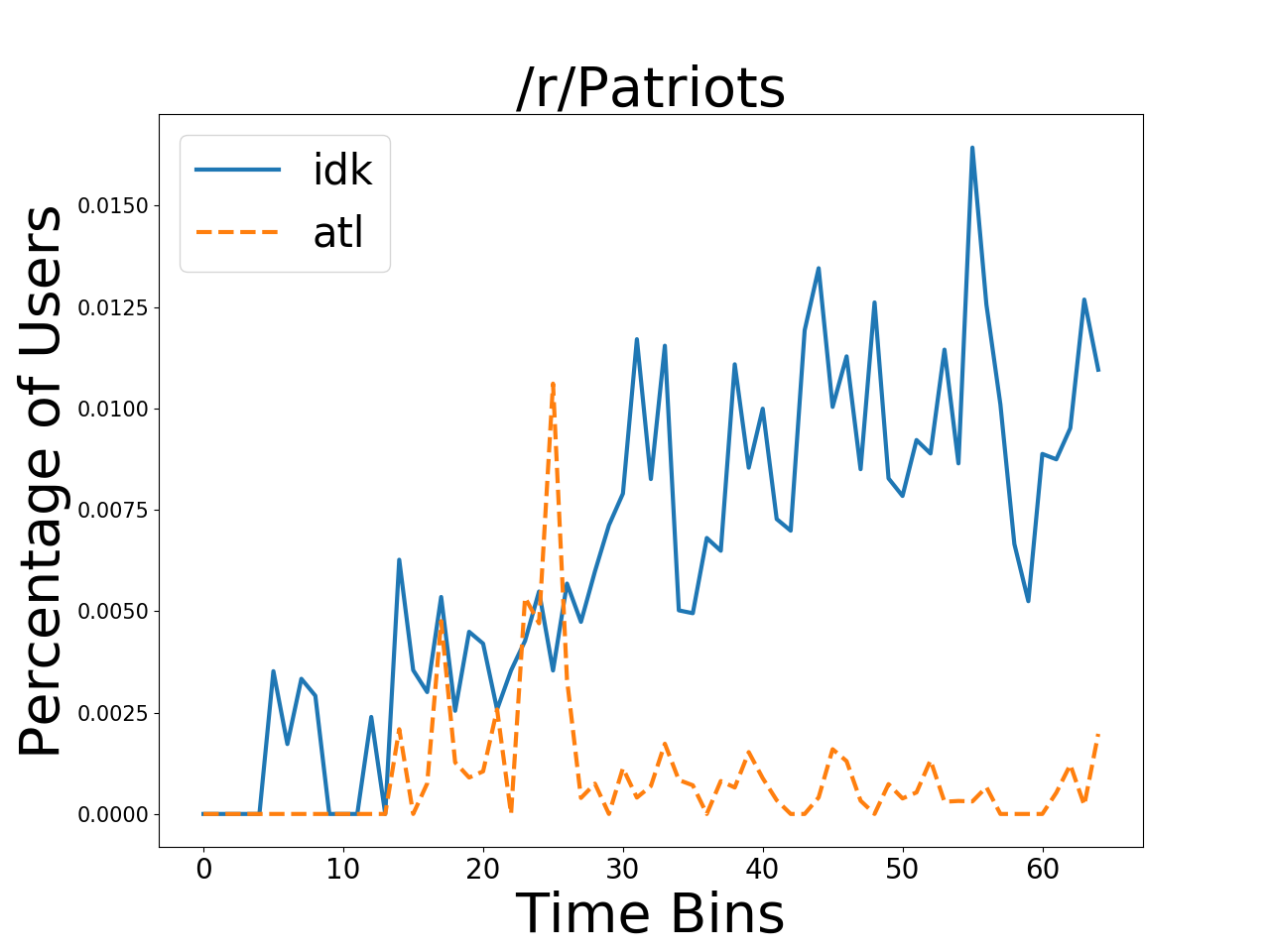}
\end{minipage}
\caption{\textbf{Left}: Distribution of the final dissemination values in /r/beer. \textbf{Center/Right}: Examples of dissemination trajectories of \textit{successful} (blue solid line) and \textit{unsuccessful} (orange dashed line) innovations.
\label{fig:zipf_dist_and_spread_trajectories}}
\vspace*{-.5cm}
\end{figure*}

Our analysis reveals some patterns common to all communities, which we present in this section. 

\subsection{Linguistic Innovations}
\label{subsec:empirical_linguisticl_innovations}

In order to explore the relative success of innovations, we consider the level of dissemination reached by an innovation to be the average dissemination value in the last six months.\footnote{This makes our measurements more robust than taking only the very last bin.} Results show that the distribution of these final dissemination values is highly skewed for all the subreddits --- see, for example, Figure~\ref{fig:zipf_dist_and_spread_trajectories} (left) for /r/beer. While a few innovations disseminate successfully (i.e., are adopted by a relatively high number of community members), most of them do not spread, and either disappear or barely appear in the last period. 


In Figure~\ref{fig:zipf_dist_and_spread_trajectories} (center/right) we show examples of successful and unsuccessful innovations. Successful innovations such as \textit{iap} (\textit{In App Purchases}) and \textit{idk} (\textit{I don't know}) show a stable increase in dissemination after their introduction, which can reach a plateau at some point (\textit{iap}) --- thus showing the  S-shaped curve typical of general processes of innovation adoption \cite{rogers2010diffusion} --- or can still be ongoing at the end of the period covered by our analysis (\textit{idk}). Unsuccessful innovations, in contrast, can either have a flat dissemination trajectory, as in the case of \textit{irl} (\textit{In Real Life}), indicating that the term has never experienced a spread in the community, or present a peak at some point, followed by a sudden decrease with no stable recovery, as for \textit{atl} (\textit{Atlanta}). 

Given these observations, we formally define the classes of \textit{successful} and \textit{unsuccessful} innovations based on the dissemination \textit{slope} of a term, computed as the difference between its average dissemination value in the first six months and in the last six months in the dissemination trajectory vector. We include in the {\em unsuccessful} class innovations with slope index $\leq$ 0, i.e., those with trajectories similar to the \textit{irl} and \textit{atl} examples in Figure~\ref{fig:zipf_dist_and_spread_trajectories} (center/right). In order to discard innovations with very low positive slope (i.e., those that do not disappear, but are only sporadically used) we only include in the {\em successful} class terms whose slope index is above the average value of the community.\footnote{Average slope index is positive for all subreddits.}
We will make use of these two classes in the prediction experiment we present in Section~\ref{sec:prediction}.


\subsection{Social Networks}
\label{subsec:empirical_social_network}



\begin{wrapfigure}{r}{7cm}\centering
\vspace*{-.5cm}
\includegraphics[width=7cm]{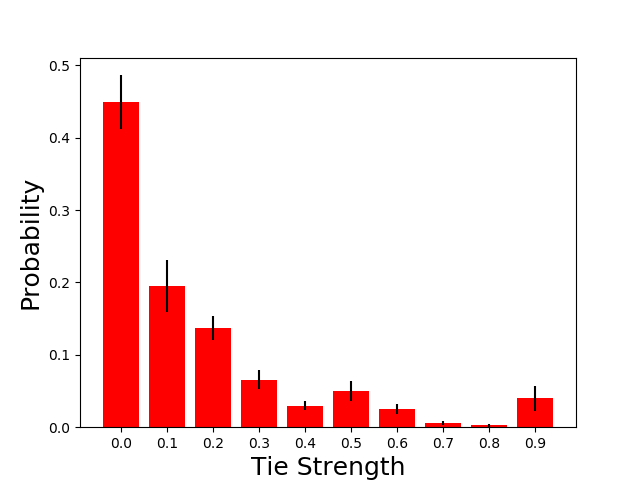}
\caption{Average tie-strength distribution for all subreddits, with standard deviation.
\label{fig:ties_distribution}}
\end{wrapfigure}

Next, we analyse the distribution of users' tie-strength values in the social graphs derived for the communities in our dataset.\footnote{A sample graph is available at \url{https://github.com/marcodel13/The-Road-to-Success}.} We find a clear pattern for all subreddits: the large majority of users have low tie-strength, with around 39\% having values $\leq$ 0.05 and almost 50\% having values $\leq$ 0.1; while, around 15 to 20\% of users have strong tie-strength, with values $\geq$ 0.5. Figure~\ref{fig:ties_distribution} shows the average tie-strength value distribution computed over all the monthly graphs of all subreddits in the dataset, with probabilities calculated for bins of size 0.1 for illustration purposes. 

%



This distribution mirrors the typical power-law distribution observed for centrality measures in online communities
\cite{mihalcea2011graph}. 
The topological properties captured by our tie-strength measure (Equation~[\ref{eq:tie-strength}]), however, are different from those captured by centrality, as already hinted at in Section~\ref{subsec:methods_social_network}. 
The three centrality measures considered (degree, betweenness, and eigenvector) correlate strongly with each other (Spearman's $r$ in range 0.85--0.89).\footnote{All correlation coefficients reported are averages across time bins and subreddits and are all significant with $p\!<\!0.05$.} But there is only a moderate correlation with tie-strength: $r$=0.63 degree, $r$=0.61 betweenness,  and $r$=0.47 eigenvector.
In addition, we observe that the three centrality measures correlate strongly with number of posts ($r$=[0.78--0.91]) in all subreddits, while we find low correlation between number of posts and tie-strength ($r$=0.31, std=0.08, $p\!<\!.05$).

These results confirm the difference between our tie-strength measure and centrality. While centrality values are \textit{global} indices of the role of a node with respect to the entire graph \cite{newman2010networks}, tie-strength captures the \textit{local} topological information around a node. In the online social communities we investigate, individuals at the core of the social network, who interact with many other individuals and have high posting activity, receive high centrality values. In contrast, high tie-strength values are the signature of users who belong to small cliques, but who do not act as hubs for the entire network.  
We take this as confirmation that our tie-strength measure does capture key features of the social structures underpinning Milroy's \shortcite{milroy1987language} theory. It remains to be seen, however, whether these social structures, as realised in large online social communities, lend support to the theory's main claims.

\section{Assessing Sociolinguistic Claims}
\label{sec:probing_theories}


In this section, we uncover the features that characterise innovators and analyse the role of strong-tie users in the dissemination process.


\subsection{Innovators}
\label{subsect:Innovators_weak_ties}


\begin{wrapfigure}{R}{7cm}\centering
\vspace*{-1cm}
\includegraphics[width=7.3cm]{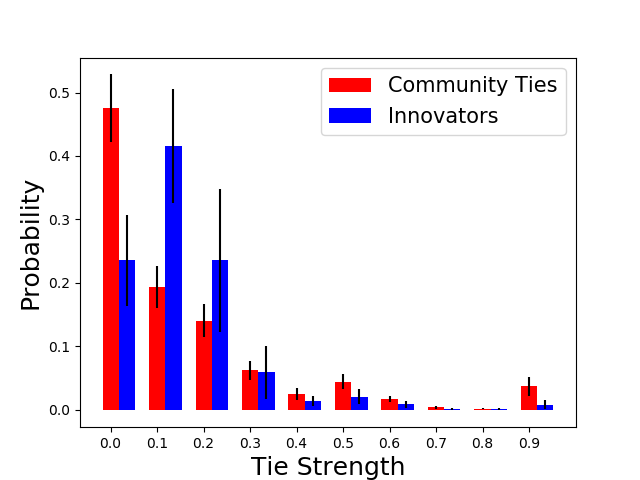}
\vspace*{-.3cm}
\caption{Comparison of the probability mass distribution of innovators' tie-strength values and avg.~values of all users in all subreddits.\label{fig:dist_first_ties}}
\end{wrapfigure}

We consider {\em innovators} those members who introduce a new term, i.e., those who use it for the very first time in a community. 
In order to verify whether innovators are weak-tie users (as hypothesised by \newcite{milroy1987language}), we compare the distribution of the tie-strength values of innovators to the general tie-strength distribution in the community. Since innovations are introduced at different points in time, the general tie-strength distribution in the community is defined as the average of the months at which innovations were introduced. 
We compute Kullback-Leibler divergence (KL), which measures how similar the behaviour of two distributions is. 
We find KL values in the range 0.15--0.4, which indicates a moderate difference between the distributions. The source of this difference can be appreciated in Figure~\ref{fig:dist_first_ties}: the tie-strength values of innovators cluster around 0.1--0.3.\footnote{The same trend holds for all subreddits. Individual plots per subreddit can be found at \url{https://github.com/marcodel13/The-Road-to-Success}.} This contrasts with the overall tie-strength distribution: innovators tend to \emph{not} be strong-tie users (lower blue than red bars for tie-strength $\geq 0.4$ in Figure~\ref{fig:dist_first_ties}) and are far less likely to have very weak tie-strength than most users   (lower blue than red bar for tie-strength $< 0.1$).

In addition to analysing tie-strength, we compare the centrality values and the posting activity of innovators and non-innovators. We observe that innovators are significantly more central than other users (for all measures considered: degree, betweenness and eigenvector) and are significantly more active in terms of number of posts than other individuals in the community (unpaired Welch's $t$-tests, all with $p\!<\!0.05$). 

Thus, a very robust pattern emerges across all subreddits, showing that innovators do have a particular profile in terms of their social standing: they do not belong to tightly connected cliques and occupy a central position in the network, as hubs with many connections of relatively low strength. 
On the one hand, this seems in line with Milroy's hypothesis, since it confirms that innovations do not arise within sub-communities that are close-knit. On the other hand, our results may also be interpreted as lending support to Labov's \shortcite{labov1972,labov2001social} characterisation of innovators as \emph{leaders}, since they occupy a core position in the network.

\subsection{Strong-Tie Users and Innovation Spread}
\label{subsect:strong_ties_innovation}



We consider strong-tie users those with a tie-strength value $\geq 0.5$. By definition, these are users who are part of cliques or sub-communities within a subreddit. As mentioned in  Section~\ref{subsec:empirical_social_network}, strong-tie users constitute between 15 and 20\% of the total number of community members. Thus, in contrast to the small scale social communities examined by Milroy and other sociolinguists, where most community members are part of some tight-knit sub-group (such as a family or a church congregation), in large  online social networks strong-tie users are a minority. Despite this, strong-tie users are not isolated in remote cliques: Our analysis shows that, in all subreddits, strong-tie users are significantly more central (with respect to all, degree, betweenness, and eigenvector centrality) and have significantly more posting activity than users with  weak tie-strength  values $\leq 0.05$, who make up the vast majority of community members --- around 39\% on average (unpaired Welch's $t$-test, $p\!<\!0.05$). Thus, in terms of centrality, strong-tie members occupy an intermediate position in the social network, forming a loose ring around the core innovators and the majority of members, who are on the periphery and are characterised by very weak tie-strength.


\begin{wrapfigure}{R}{7cm}\centering
\vspace*{-1cm}\hspace*{-.3cm}
\includegraphics[width=7.7cm]{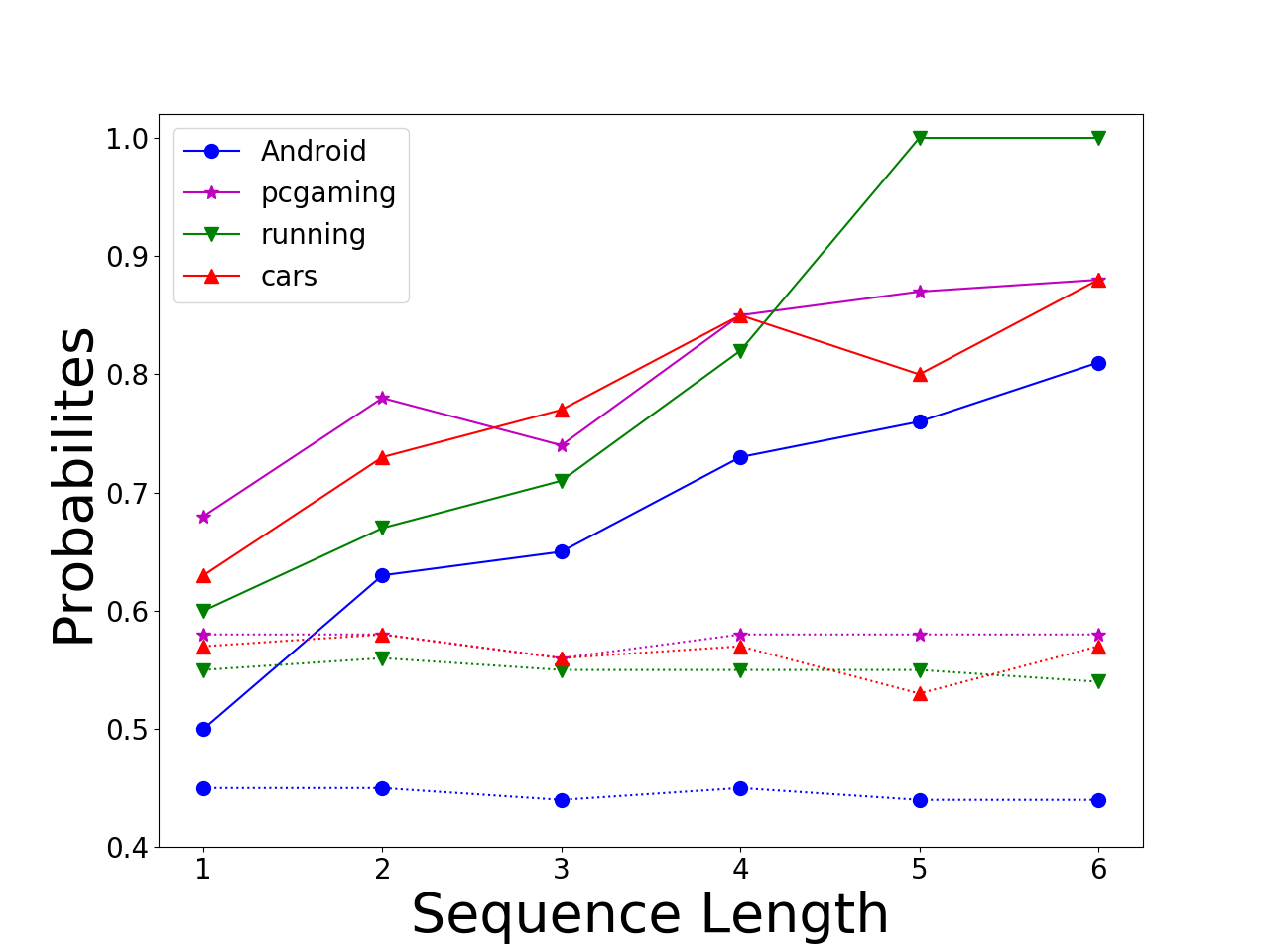}
\vspace*{-.6cm}
\caption{Probability of dissemination increase after a term is adopted by a strong-tie user (solid line) or by a weak-tie user (dotted line) for $k$ consecutive months, computed for $k$ in range $[1-6]$.\label{fig:adopters_strong_1}}
\vspace*{-.2cm}
\end{wrapfigure}
To analyse the role of strong-tie users in the innovation diffusion process, we proceed as follows: We identify any time $t_i$ in the tie-strength trajectory vector when the innovation is used by some strong-tie member for $k$ consecutive months. We then check its average dissemination in the period up to time $t_i$ and compare it to the average dissemination in the six months following $t_{i+k-1}$. We find that when $k=1$ (i.e., when an innovation has been used by a strong-tie member only in one month) the probability that the dissemination increases in the next six months is around 50\% for all subreddits --- a value similar to the likelihood of dissemination increase after the usage by a weak-tie user. However, as $k$ increases, and thus the adoption by strong-tie users becomes more stable, a future increase in dissemination becomes progressively more likely, for all the subreddits. Importantly, the same effect is not observed for weak-tie users, for whom, independently from the number of months, the probability of a future increase in dissemination is always approximately the same. Figure~\ref{fig:adopters_strong_1} shows examples of how the probability of dissemination changes after the adoption by either strong- or weak-tie users (see Appendix A for full results).




These results, thus, are consistent with Milroy's \shortcite{milroy1987language} claims, and furthermore show that innovation diffusion is connected to {\em sustained} adoption by strong-tie community members.

\section{Predicting Innovation Success}
\label{sec:prediction}


Most innovations do not succeed in becoming community norms, but some do. Here we assess whether information about the tie-strength of members who use an innovation in the first months after its introduction can predict whether it will be successful in the future.  This provides further theoretical insight into the importance of tie-strength for innovation diffusion, and has practical significance by contributing to identifying new terms that NLP systems should be able to process.
Our aim here is not to maximise prediction accuracy---which is likely to require taking into account several factors beyond users' tie-strength---but rather to explore whether the statistical effects we have uncovered are strong enough to have some predictive power.

We approach this as a binary classification task, making use of the distinction between \textit{successful} and \textit{unsuccessful} innovations defined in Section~\ref{sec:empirical_observations}.
We extract a subvector of length $k$ from the tie-strength trajectory vector of innovations and use it as features for the prediction. For instance, with $k=3$, we use the tie-strength information from the first three months of usage of a term to predict if it will be successful or not, leveraging subvectors of increasing magnitudes. 
We use Python's  \texttt{scikit-learn} Random Forest classifier with default parameters\footnote{\url{http://scikit-learn.org/stable/modules/generated/sklearn.ensemble.RandomForestClassifier.html}.}
and perform 100-run cross-validation, using 90\% of the data for training and 10\% for testing. We compare our results against a {\em weighted baseline}, whereby the two labels (successful/unsuccessful) are randomly assigned taking into account their frequency in the training set. The classes are fairly balanced across subreddits, with an average proportion of 55\% successful  and 45\% unsuccessful.

When leveraging tie-strength information from only the first 3 months of usage, we obtain F1 results that are significantly higher than the baseline for 12 out the 20 subreddits. But overall, performance remains rather low, with an average F1-score for the {\em successful} class of 0.62 vs.~0.58 for the baseline. Given that new terms are introduced by users with relatively low tie-strength (as shown in Section~\ref{subsect:Innovators_weak_ties}), arguably in the initial few months before a novel term is picked up by a strong-tie user, there is little difference between successful and unsuccessful innovations. With tie-strength information from the first 6 months of usage, we are able to make predictions with results significantly above baseline for 18 out 20 subreddits, with an average F1-score of 0.68. Not surprisingly, performance increases substantially when information for a longer period (first/second year of usage) is exploited, reaching an average F1-score of 0.76, significantly above the baseline for all communities. Detailed results, including precision, recall, and F1-score for each subreddit, can be found in Appendix B. In Figure~\ref{fig:plots_classifcation} we graphically illustrate the results for a few subreddits, which are representative of the general trend observed.

\begin{figure}[h] \centering
 \includegraphics[width=8cm]{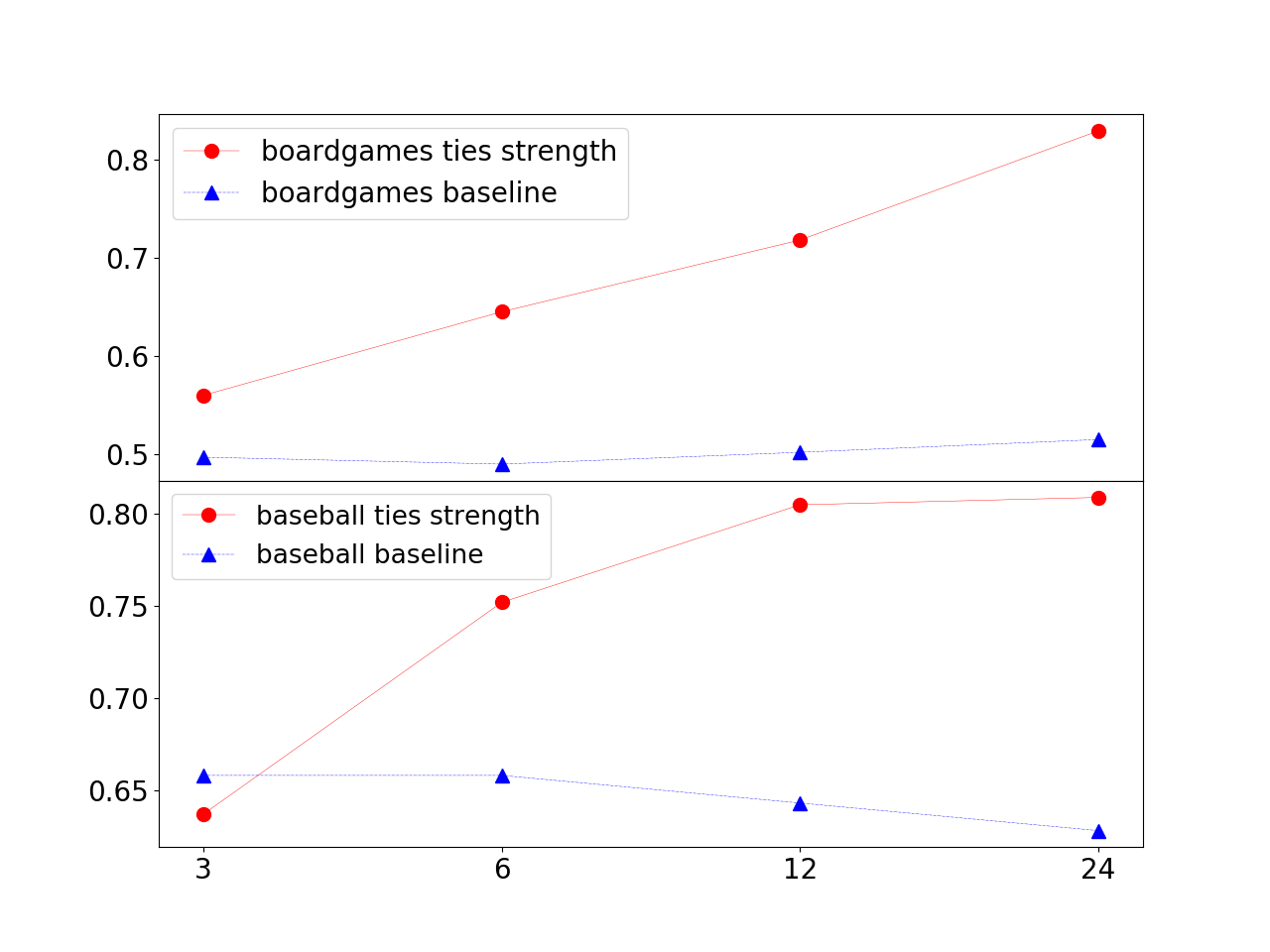} 
\hspace*{-.5cm}
 \includegraphics[width=8cm]{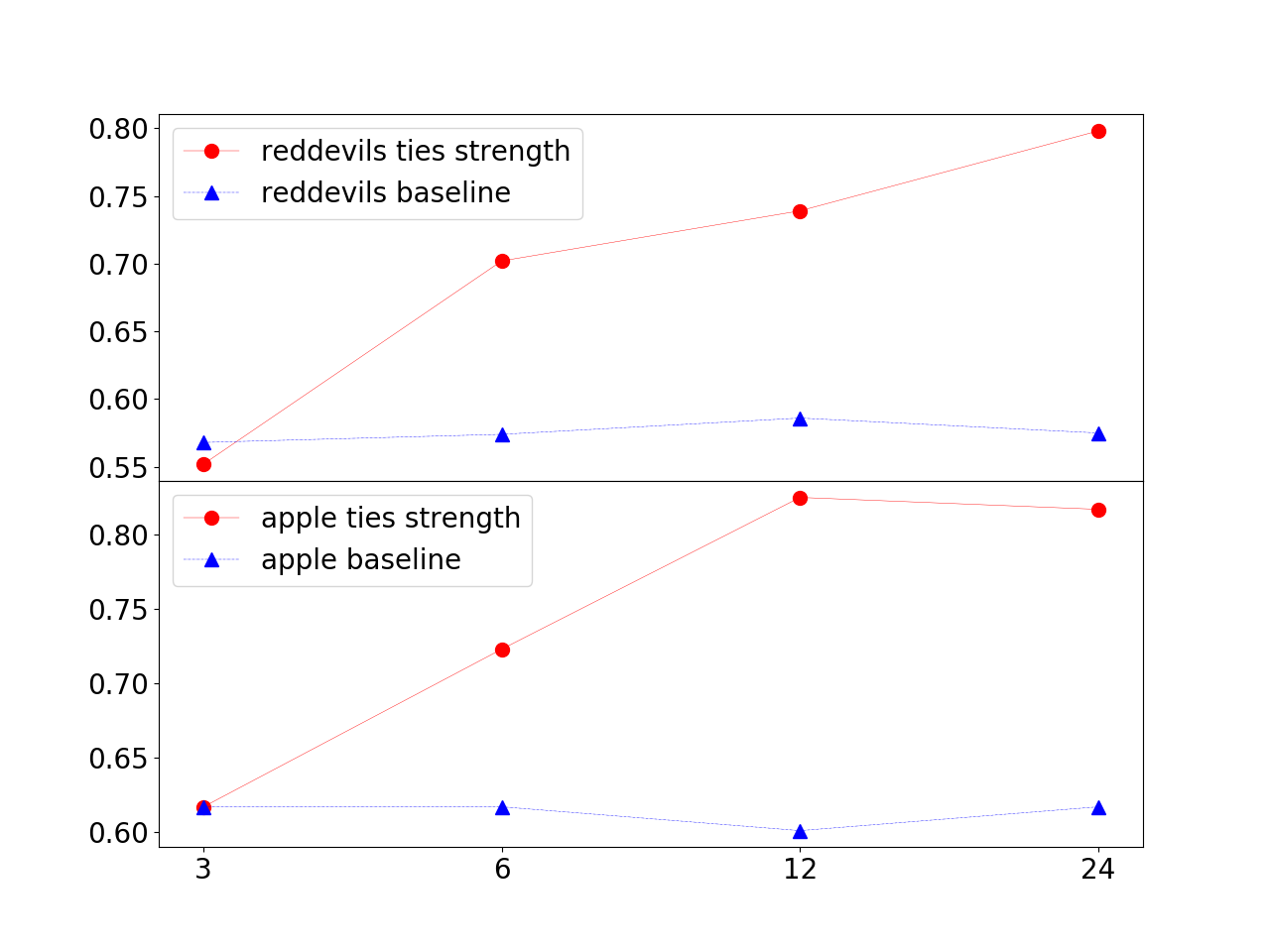}
 \caption{F-score (y-axis) for the successful class obtained with the ties-strength values of the first $k$ months (x-axis) after the introduction of a term.}
    \label{fig:plots_classifcation}
\end{figure}

\clearpage

\section{Conclusions}
\label{sec:discussion_conclusion}


 \vspace*{-3pt}
This work has provided a large-scale analysis of the interplay between the birth and spread of new terms and users' social standing in large online social communities. Building on sociolinguistic theories --- in particular, Milroy's \shortcite{milroy1987language} version of {\em The Strength of Weak Ties} theory --- we have proposed a simple measure to quantify tie-strength in Milroy's sense and have used it in combination with common centrality measures to uncover the characteristics of innovators and to assess the role of strong-tie users in the dissemination process. 

Regarding innovators, our results show that they are central community members, connected to many other users but with relatively low tie-strength. As for strong-tie users, we find that in online social networks they are a small proportion of community members, organised in small cliques, and that they do play an important role in the spread of an innovation, presumably by spreading new terms introduced by innovators into their sub-groups. 
The patterns we have revealed are surprisingly consistent across the 20 online communities we have investigated. 

Our work opens a range of interesting questions. In particular, we are looking forward to investigating how the patterns we have observed are affected by users' membership in multiple forums and how these multi-community users transfer new terms between communities.


\section*{Acknowledgements}

This research has received funding from the Netherlands Organisation for Scientific Research (NWO) under VIDI grant nr.~276-89-008, {\em Asymmetry in Conversation}. We thank the anonymous reviewers for their comments as well as the area chairs and PC chairs of COLING 2018. 

\bibliography{innovations}
\bibliographystyle{acl}


\appendix

\clearpage

\section*{Appendix A: Probability of Dissemination Increase}

The table in this appendix complements the results presented in Section \ref{subsect:strong_ties_innovation}. 

\begin{table}[h]\small \centering
\begin{tabular}{|l|c@{\ \ }c|c@{\ \ }c|c@{\ \ }c|c@{\ \ }c|c@{\ \ }c|c@{\ \ }c|}
\hline
 & \multicolumn{2}{c|}{$k$=1} & \multicolumn{2}{c|}{$k$=2 } & \multicolumn{2}{c|}{$k$= 3} & \multicolumn{2}{c|}{$k$=4} & \multicolumn{2}{c|}{$k$=5} & \multicolumn{2}{c|}{$k$=6}
\\\hline
\bf subreddit& \bf strong & \bf weak & \bf strong & \bf weak & \bf strong & \bf weak & \bf strong & \bf weak & \bf strong & \bf weak & \bf strong & \bf weak
\\\hline
r/Android &0.5 & 0.45 &0.63 & 0.45 &0.65 & 0.44 &0.73 & 0.45 &0.76 & 0.44 &0.81 & 0.44\\
r/apple &0.58 & 0.47 &0.7 & 0.48 &0.73 & 0.48 &0.8 & 0.47 &0.8 & 0.47 &0.81 & 0.47\\
r/baseball &0.62 & 0.55 &0.73 & 0.55 &0.71 & 0.54 &0.76 & 0.56 &0.75 & 0.55 &0.75 & 0.54 \\
r/beer &0.51 & 0.49 &0.62 & 0.49 &0.68 & 0.49 &0.69 & 0.48 &0.72 & 0.48 &0.73 & 0.48\\
r/boardgames &0.71 & 0.6 &0.83 & 0.6 &0.8 & 0.58 &0.88 & 0.6 &0.7 & 0.56 &0.9 & 0.6 \\
r/cars &0.63 & 0.57 &0.73 & 0.58 &0.77 & 0.56 &0.85 & 0.57 &0.8 & 0.53 &0.88 & 0.57\\
r/FinalFantasy &0.61 & 0.58 &0.59 & 0.57 &1.0 & 0.56 & -- & -- & -- & -- & -- & -- \\
r/Guitar &0.57 & 0.53 &0.7 & 0.54 &0.7 & 0.52 &0.81 & 0.53 &0.89 & 0.53 &0.77 & 0.53\\
r/harrypotter &0.53 & 0.51 &0.5 & 0.5 &1.0 & 0.51 & -- & -- & -- & -- & -- & -- \\
r/hockey &0.74 & 0.61 &0.8 & 0.61 &0.89 & 0.62 &0.76 & 0.58 &0.83 & 0.61 &0.94 & 0.62\\
r/Liverpool &0.61 & 0.52 &0.61 & 0.51 &0.67 & 0.53 &0.67 & 0.49 & -- & -- & -- & -- \\
r/Patriots &0.65 & 0.64 &0.62 & 0.63 &0.6 & 0.65 &0.67 & 0.65 & -- & -- & -- & -- \\
r/pcgaming &0.68 & 0.58 &0.78 & 0.58 &0.74 & 0.56 &0.85 & 0.58 &0.87 & 0.58 &0.88 & 0.58 \\
r/photography &0.65 & 0.57 &0.73 & 0.57 &0.82 & 0.57 &0.84 & 0.58 &0.75 & 0.57 &0.73 & 0.57\\
r/pokemon &0.54 & 0.48 &0.66 & 0.48 &0.69 & 0.48 &0.69 & 0.48 &0.71 & 0.47 &0.71 & 0.47\\
r/poker &0.57 & 0.57 &0.7 & 0.57 &1.0 & 0.57 & -- & -- & -- & -- & -- & - \\
r/reddevils &0.56 & 0.54 &0.54 & 0.53 &0.6 & 0.52 &0.6 & 0.5 &0.4 & 0.5 &1.0 & 0.49 \\
r/running &0.6 & 0.55 &0.67 & 0.56 &0.71 & 0.55 &0.82 & 0.55 &1.0 & 0.55 &1.0 & 0.54\\
r/StarWars &0.61 & 0.5 &0.77 & 0.51 &0.82 & 0.51 &0.88 & 0.5 &0.9 & 0.5 &0.91 & 0.5\\
r/subaru &0.49 & 0.53 &0.53 & 0.51 &0.92 & 0.52 &0.8 & 0.48 & -- & -- & -- & -- \\
\hline
\end{tabular}
\caption{Probability of increase in dissemination of a linguistic innovation after being used by a \textbf{strong-tie} or a \textbf{weak-tie} user for $k$ consecutive months. Missing values indicate no such condition was found in a community.
}
\end{table}

\section*{Appendix B: Detailed Results on Success Prediction}

The following table gives detailed results for the prediction task described in Section~\ref{sec:prediction}. Statistical significance between tie-strength features and the weighted baseline is computed with the dependent $t$-test for paired samples implemented by Python's \texttt{scipy} package on results from 100 runs. 

\begin{table}[h] \centering \small
\begin{tabular}{|@{\ }l@{\ }|c@{\ \ }|ccc|ccc|ccc|ccc@{\ }|}\hline
& & \multicolumn{3}{c|}{$k$=3} & \multicolumn{3}{c|}{$k$=6} & \multicolumn{3}{c|}{$k$=12} & \multicolumn{3}{c|}{$k$=24}\\\hline
\bf subreddit  & & \bf P &  \bf R &  \bf F1  &  \bf P & \bf  R &  \bf F1  &  \bf P &  \bf R &  \bf F1  &  \bf P &  \bf R &  \bf F1  \\   \hline\hline

Android & \bf t & 0.49 & 0.5 & 0.49 & 0.55 & 0.57 & 0.56 & 0.59 & 0.59 & 0.59 & 0.68 & 0.69 & 0.68\\
 		     & \bf b & 0.37 & 0.38 & 0.37 & 0.38 & 0.38 & 0.38 & 0.4 & 0.41 & 0.4 & 0.39 & 0.39 & 0.39\\ \hline
apple & \bf t & 0.64$^\#$  & 0.6$^\#$  & 0.62$^\#$  & 0.72 & 0.73 & 0.72 & 0.8 & 0.85 & 0.82 & 0.78 & 0.86 & 0.82\\
    & \bf b & 0.64$^\#$  & 0.62$^\#$  & 0.63$^\#$  & 0.66 & 0.6 & 0.63 & 0.62 & 0.61 & 0.61 & 0.65 & 0.61 & 0.63\\ \hline
baseball & \bf t & 0.63 & 0.65$^\#$  & 0.64$^\#$  & 0.71 & 0.79 & 0.75 & 0.73 & 0.89 & 0.8 & 0.74 & 0.89 & 0.81\\
    & \bf b  & 0.67 & 0.67$^\#$  & 0.67$^\#$  & 0.66 & 0.68 & 0.67 & 0.64 & 0.67 & 0.65 & 0.63 & 0.65 & 0.64\\ \hline
beer & \bf t & 0.52$^\#$  & 0.52$^\#$  & 0.52 $^\#$ & 0.58 & 0.62 & 0.6 & 0.59 & 0.65 & 0.62 & 0.64 & 0.7 & 0.67\\
    & \bf b & 0.52$^\#$  & 0.51$^\#$  & 0.51$^\#$  & 0.51 & 0.49 & 0.5 & 0.49 & 0.5 & 0.49 & 0.52 & 0.5 & 0.51\\ \hline
boardgames & \bf t & 0.56$^\#$  & 0.56$^\#$  & 0.56 & 0.64 & 0.65 & 0.64 & 0.69 & 0.74 & 0.71 & 0.81 & 0.85 & 0.83\\
   				    & \bf b & 0.52$^\#$  & 0.51$^\#$  & 0.51 & 0.5 & 0.51 & 0.5 & 0.55 & 0.5 & 0.52 & 0.54 & 0.52 & 0.53\\ \hline
cars & \bf t & 0.66 & 0.66 & 0.66 & 0.7 & 0.76 & 0.73 & 0.71 & 0.82 & 0.76 & 0.72 & 0.85 & 0.78\\
    & \bf b & 0.6 & 0.61 & 0.6 & 0.6 & 0.61 & 0.6 & 0.6 & 0.6 & 0.6 & 0.58 & 0.59 & 0.58\\ \hline
Finalfantasy & \bf t & 0.67$^\#$  & 0.6$^\#$  & 0.63$^\#$  & 0.65$^\#$  & 0.7 & 0.67 & 0.76 & 0.84 & 0.8 & 0.79 & 0.86 & 0.82\\
   				   & \bf b & 0.65$^\#$  & 0.58$^\#$  & 0.61$^\#$  & 0.63$^\#$  & 0.58 & 0.6 & 0.63 & 0.6 & 0.61 & 0.64 & 0.6 & 0.62\\ \hline
Guitar & \bf t & 0.68 & 0.65 & 0.66 & 0.68 & 0.76 & 0.72 & 0.71 & 0.81 & 0.76 & 0.75 & 0.85 & 0.8\\
   		  & \bf b & 0.58 & 0.6 & 0.59 & 0.58 & 0.61 & 0.59 & 0.56 & 0.6 & 0.58 & 0.58 & 0.6 & 0.59\\ \hline
harrypotter & \bf t & 0.53$^\#$  & 0.55$^\#$  & 0.54 & 0.57 & 0.58 & 0.57 & 0.56 & 0.6 & 0.58 & 0.51 & 0.55 & 0.53\\
  				   & \bf b & 0.49$^\#$  & 0.51$^\#$  & 0.5 & 0.47 & 0.48 & 0.47 & 0.48 & 0.52 & 0.5 & 0.45 & 0.48 & 0.46\\ \hline
hockey & \bf t & 0.72 & 0.75 & 0.73 & 0.72 & 0.79 & 0.75 & 0.68$^\#$  & 0.81 & 0.74 & 0.74 & 0.88 & 0.8\\
  		    & \bf b & 0.64 & 0.64 & 0.64 & 0.68 & 0.62 & 0.65 & 0.66$^\#$  & 0.63 & 0.64 & 0.64 & 0.62 & 0.63\\ \hline
Liverpool & \bf t & 0.59 & 0.59 & 0.59 & 0.62 & 0.64 & 0.63 & 0.58$^\#$  & 0.61 & 0.59 & 0.69 & 0.78 & 0.73\\
    		  & \bf b & 0.53 & 0.5 & 0.51 & 0.53 & 0.52 & 0.52 & 0.54$^\#$  & 0.51 & 0.52 & 0.53 & 0.53 & 0.53\\ \hline
Patriots & \bf t & 0.72$^\#$  & 0.74 & 0.73 & 0.8 & 0.81 & 0.8 & 0.81 & 0.88 & 0.84 & 0.81 & 0.92 & 0.86\\
    		 & \bf b & 0.71$^\#$  & 0.69 & 0.7 & 0.64 & 0.69 & 0.66 & 0.67 & 0.71 & 0.69 & 0.65 & 0.69 & 0.67\\ \hline
pcgaming & \bf t & 0.78 & 0.7$^\#$  & 0.74 & 0.77 & 0.83 & 0.8 & 0.82 & 0.86 & 0.84 & 0.79 & 0.89 & 0.84\\
    		  & \bf b & 0.66 & 0.66$^\#$  & 0.66 & 0.68 & 0.68 & 0.68 & 0.67 & 0.68 & 0.67 & 0.68 & 0.7 & 0.69\\ \hline
photography & \bf t & 0.72 & 0.64 & 0.68 & 0.74 & 0.72 & 0.73 & 0.74 & 0.8 & 0.77 & 0.77 & 0.85 & 0.81\\
   				    & \bf b & 0.63 & 0.62 & 0.62 & 0.65 & 0.61 & 0.63 & 0.64 & 0.61 & 0.62 & 0.63 & 0.61 & 0.62\\ \hline
pokemon & \bf t & 0.65 & 0.67 & 0.66 & 0.58 & 0.66 & 0.62 & 0.62 & 0.77 & 0.69 & 0.68 & 0.79 & 0.73\\
   			   & \bf b & 0.53 & 0.52 & 0.52 & 0.51 & 0.5 & 0.5 & 0.53 & 0.52 & 0.52 & 0.52 & 0.49 & 0.5\\ \hline
poker & \bf t & 0.58$^\#$  & 0.62$^\#$  & 0.6$^\#$  & 0.6$^\#$  & 0.64$^\#$  & 0.62$^\#$  & 0.58 & 0.63$^\#$  & 0.6$^\#$  & 0.65 & 0.78 & 0.71\\
   		  & \bf b & 0.62$^\#$  & 0.6$^\#$  & 0.61$^\#$  & 0.61$^\#$  & 0.59$^\#$  & 0.6 $^\#$ & 0.63 & 0.59$^\#$  & 0.61$^\#$  & 0.58 & 0.6 & 0.59\\ \hline
reddevils & \bf t & 0.57$^\#$  & 0.54$^\#$  & 0.55$^\#$  & 0.68 & 0.72 & 0.7 & 0.7 & 0.78 & 0.74 & 0.74 & 0.86 & 0.8\\
  			   & \bf b & 0.59$^\#$  & 0.58$^\#$  & 0.58$^\#$  & 0.58 & 0.59 & 0.58 & 0.58 & 0.61 & 0.59 & 0.59 & 0.59 & 0.59\\ \hline
running & \bf t & 0.58$^\#$  & 0.55$^\#$  & 0.56$^\#$  & 0.64 & 0.68 & 0.66 & 0.65 & 0.77 & 0.7 & 0.66 & 0.81 & 0.73\\
   		    & \bf b & 0.57$^\#$  & 0.58$^\#$  & 0.57$^\#$  & 0.58 & 0.61 & 0.59 & 0.56 & 0.61 & 0.58 & 0.58 & 0.58 & 0.58\\ \hline
Starwars & \bf t & 0.63 & 0.63$^\#$  & 0.63$^\#$  & 0.57$^\#$  & 0.63$^\#$  & 0.6$^\#$  & 0.6$^\#$  & 0.67 & 0.63 & 0.65 & 0.76 & 0.7\\
   			   & \bf b & 0.55 & 0.58$^\#$  & 0.56$^\#$  & 0.6$^\#$  & 0.61$^\#$  & 0.6$^\#$  & 0.56$^\#$  & 0.59 & 0.57 & 0.6 & 0.59 & 0.59\\ \hline
subaru & \bf t & 0.69 & 0.64$^\#$  & 0.66 & 0.69 & 0.79 & 0.74 & 0.69 & 0.83 & 0.75 & 0.69 & 0.84 & 0.76\\
   		   & \bf b & 0.63 & 0.61$^\#$  & 0.62 & 0.61 & 0.61 & 0.61 & 0.61 & 0.61 & 0.61 & 0.62 & 0.59 & 0.6\\ \hline\hline

 \bf Average & \bf t & 0.63 & 0.62 & 0.62 & 0.7 & 0.66  & 0.68  & 0.68 & 0.76  & 0.72  &  0.71 & 0.81 & 0.76  \\
     			& \bf b & 0.58 & 0.57 & 0.58  & 0.58  & 0.57  & 0.58  & 0.58  & 0.58  &  0.58 & 0.58 & 0.58 & 0.58  \\ \hline
\end{tabular}
\caption{Average precision (P), recall (R), and F1-score for the {\em successful} class over 100 runs with 10-fold cross-validation. $k$= length of the tie-strength vector used for the prediction (corresponding to number of months); \textbf{t} / \textbf{b}= results obtained using tie-strength information and the weighted baseline, respectively. Difference between \textbf{t} and \textbf{b} is significant ($ p <   0.05$) except when marked with $^\#$.}
\end{table}

\end{document}